\def\year{2022}\relax
\documentclass[letterpaper]{article} % DO NOT CHANGE THIS
\usepackage{aaai22}  % DO NOT CHANGE THIS
\usepackage{times}  % DO NOT CHANGE THIS
\usepackage{helvet}  % DO NOT CHANGE THIS
\usepackage{courier}  % DO NOT CHANGE THIS
\usepackage[hyphens]{url}  % DO NOT CHANGE THIS
\usepackage{graphicx} % DO NOT CHANGE THIS
\usepackage{natbib}  % DO NOT CHANGE THIS AND DO NOT ADD ANY OPTIONS TO IT
\usepackage{caption} % DO NOT CHANGE THIS AND DO NOT ADD ANY OPTIONS TO IT
\DeclareCaptionStyle{ruled}{labelfont=normalfont,labelsep=colon,strut=off} % DO NOT CHANGE THIS
\usepackage{algorithm}
\usepackage{algorithmic}
\usepackage{amsfonts}
\usepackage{amsmath}
\usepackage{mathtools}

\usepackage{multirow}
\usepackage{booktabs}

\usepackage{newfloat}
\usepackage{listings}
\floatstyle{ruled}
\newfloat{listing}{tb}{lst}{}
\floatname{listing}{Listing}

\title{Adaptive Kernel Graph Neural Network}
\author {
    Mingxuan Ju\textsuperscript{1,2},
    Shifu Hou\textsuperscript{2},
    Yujie Fan\textsuperscript{2},
    Jianan Zhao\textsuperscript{1,2},
    Yanfang Ye\textsuperscript{1,2}\thanks{Corresponding Author.},
    Liang Zhao\textsuperscript{3}
    \\
}
\affiliations {
    \textsuperscript{1} University of Notre Dame, Notre Dame, IN 46556 \\
    \textsuperscript{2} Case Western Reserve University, Cleveland, OH 44106 \\
    \textsuperscript{3} Emory University, Atlanta, GA 30322 \\
    \{mju2, jzhao8, yye7\}@nd.edu, \{sxh1055,yxf370\}@case.edu, liang.zhao@emory.edu

}

\begin{document}

\maketitle

\begin{abstract}
Graph neural networks (GNNs) have demonstrated great success in representation learning for graph-structured data. The layer-wise graph convolution in GNNs is shown to be powerful at capturing graph topology. During this process, GNNs are usually guided by pre-defined kernels such as Laplacian matrix, adjacency matrix, or their variants. However, the adoptions of pre-defined kernels may restrain the generalities to different graphs: mismatch between graph and kernel would entail sub-optimal performance. For example, GNNs that focus on low-frequency information may not achieve satisfactory performance when high-frequency information is significant for the graphs, and vice versa. To solve this problem, in this paper, we propose a novel framework - i.e., namely Adaptive Kernel Graph Neural Network (AKGNN) - which learns to adapt to the optimal graph kernel in a unified manner at the first attempt. In the proposed AKGNN, we first design a data-driven graph kernel learning mechanism, which adaptively modulates the balance between all-pass and low-pass filters by modifying the maximal eigenvalue of the graph Laplacian. Through this process, AKGNN learns the optimal threshold between high and low frequency signals to relieve the generality problem. Later, we further reduce the number of parameters by a parameterization trick and enhance the expressive power by a global readout function. Extensive experiments are conducted on acknowledged benchmark datasets and promising results demonstrate the outstanding performance of our proposed AKGNN by comparison with state-of-the-art GNNs. The source code is publicly available at: \url{https://github.com/jumxglhf/AKGNN}.
\end{abstract}

\section{Introduction}
\label{intro}
Graph-structured data have become ubiquitous in the real world, such as social networks, knowledge graphs, and molecule structures. Mining and learning expressive node representations on these graphs can contribute to a variety of realistic challenges and applications. The emphasis of this work lies in the node representation learning on graphs, aiming to generate node embeddings that are expressive with respect to downstream tasks such as node classification \cite{kipf2016semi,klicpera2019predict,velivckovic2017graph}. Current state-of-the-art frameworks could be categorized as graph convolutions, where nodes aggregate information from their direct neighbors with fixed guiding kernels, e.g., different versions of adjacency or Laplacian matrices. And information of high-order neighbors could be captured in an iterative manner by stacked convolution layers.

Although the results are promising, recent researches have shown that such a propagation mechanism entails certain challenges. Firstly, \cite{chen2020measuring,oono2019graph} illustrate that the performance of GNNs could be deteriorated by over-smoothing if excessive layers are stacked. As proved in the work of \cite{zhu2021interpreting}, graph convolution could be summarized as a case of Laplacian smoothing, in which adjacent nodes become inseparable after multiple layers. \cite{oono2019graph} shows that multiple non-linearity functions between stacked convolution layers further antagonize this problem. Moreover, these aforementioned propagation layers cause the over-fitting problem \cite{wang2019improving}. In current GNNs, each layer serves as a parameterized filter where graph signals are first amplified or diminished and then combined. Adding more layers aims to capture high-order information beneficial for the downstream classification but it meanwhile introduces the number of trainable parameters, which might cancel out the intended benefits as real-world data is often scarcely labeled \cite{zhao2020data,chen2020measuring}. Effective frameworks such as JK-Nets \cite{xu2018representation} and DAGNN \cite{liu2020towards} overcome the over-smoothing issue by a global readout function between propagation layers, making local information from early layers directly accessible during the inference phase. And the over-fitting issue is conquered by a single learnable matrix, placed before all propagation layers, to approximate parameters of all layers \cite{wu2019simplifying}.

Another far less researched issue rests in the fixed graph kernels (e.g., adjacency matrix, Laplacian matrix, or their variants) that current GNNs model on, which restricts their generalization to different graphs. \cite{ma2020unified,zhu2021interpreting,dong2016learning} prove that the current GNNs could be explained in a unified framework, where the output node representation minimizes two terms: 1) the distances between adjacent nodes and 2) the similarities between the input and output signals. Some GNNs such as GCN \cite{kipf2016semi} mainly focus on the latter term, which solely extracts low-frequency information. Others like APPNP \cite{klicpera2019predict} merges these two terms by introducing original signals through teleport connection after low-pass filtering, and hence brings a certain degree of high-frequency signals. But in reality, it is difficult to determine what and how much information should be encoded, unless experiments are conducted across algorithms with different hyperparameters. Merely considering one kind of information might not satisfy the needs of various downstream tasks, while introducing extra information could jeopardize the decision boundary. Some very recent works such as GNN-LF and GNN-HF \cite{zhu2021interpreting} utilize models with different pre-defined graph kernels to adapt to various datasets. These models either focus on high or low frequency signals. Still, experiments need to be conducted on both in order to know which one works out best, and the threshold between low and high frequency signals needs to be defined via human experts, which might be sub-optimal under some circumstances. To the best of our knowledge, there has not yet a unified framework that solves this problem. To fill this research gap, in this paper, we propose Adaptive Kernel Graph Neural Network (i.e., AKGNN) where a novel adaptive kernel mechanism is devised to self-learn the optimal threshold between high and low frequency signals for downstream tasks.

Specifically, to effectively combat the generality issue entailed by fixed graph kernel, AKGNN dynamically adjusts the maximal eigenvalue of graph Laplacian matrix at each layer such that the balance between all-pass and low-pass filters is dynamically optimized. And through this process, the optimal trade-off between high and low frequency signals is learned. From the spatial point of view, our model is able to raise the weight of self-loop when neighbors are not informative (i.e., all-pass filter from spectral view), or focus more on adjacent nodes when neighbors are helpful (i.e., low-pass filter). To prevent the over-fitting problem, we modify the parameterization trick proposed in \cite{wu2019simplifying}, wherein learnable parameters of all convolution layers, except maximal eigenvalues, are compressed and approximated by a single matrix. Nevertheless, it is possible that different nodes require information from neighbors of distinct orders and hence we utilize a global readout function \cite{xu2018representation} to capture node embeddings at different orders. 

Finally, we demonstrate the legitimacy of the proposed AKGNN through theoretical analysis and empirical studies, where it is able to achieve state-of-the-art results on node classification at acknowledged graph node classification benchmark datasets, and persistently retain outstanding performance even with an exaggerated amount of convolution layers across all benchmark datasets.

\section{Problem and Related Work}

Let $G = (V,E)$ denote a graph, in which $V$ is the set of $|V| = N$ nodes and $E \subseteq V \times V$ is the set of $|E|$ edges between nodes. Adjacency matrix is denoted as $\textbf{A} \subseteq \{0,1\}^{N \times N}$. The element $a_{ij}$ in $i$-th row and $j$-th column of $\textbf{A}$ equals to 1 if there exists an edge between nodes $v_i$ and $v_j$ or equals to 0 otherwise. Laplacian matrix of a graph is denoted as $\mathbf{L} = \mathbf{D} - \mathbf{A}$ or its normalized form $\mathbf{L} = \mathbf{D}^{-\frac{1}{2}}(\mathbf{D} - \mathbf{A})\mathbf{D}^{-\frac{1}{2}} = \mathbf{I} - \mathbf{D}^{-\frac{1}{2}}\mathbf{A}\mathbf{D}^{-\frac{1}{2}}$, where $\mathbf{D}$ is the diagonal degree matrix $\mathbf{D}$ = $diag(d(v1),\dots,d(v_N))$ and $\mathbf{I}$ is the identity matrix. Spectral graph theories have studied the properties of a graph by analyzing the eigenvalues and eigenvectors of $\mathbf{L}$ \cite{kipf2016semi,defferrard2016convolutional}, and our model adaptively modifies the maximal eigenvalue of $\mathbf{L}$ to learn the optimal trade-off between all-pass and low-pass filters.

\noindent \textbf{Node Classification on Graphs.} We focus on  node classification on graph. $\mathbf{X} \in \mathbb{R}^{N \times d}$ represents the feature matrix of a graph, where node $v_i$ is given with a feature vector $\mathbf{x}_i \in \mathbb{R}^d$ and $d$ is the dimension size. $\mathbf{Y} \subseteq \{0,1\}^{N \times C}$ denotes the label matrix of a graph, where $C$ is the number of total classes. Given $M$ labeled nodes ($0 < M \ll N$) with label $\mathbf{Y}^L$ and $N-M$ unlabeled nodes with missing label $\mathbf{Y}^U$, the objective of node classification is learning a function $f:G, \mathbf{X}, \mathbf{Y}^L \rightarrow \mathbf{Y}^U$ to predict missing labels $\mathbf{Y}^U$. Traditional solutions to this problem are mainly based on Deepwalk \cite{ando2007learning,pang2017graph,dong2016learning}. Recently GNNs have emerged as a class of powerful approaches for this problem. GCN \cite{kipf2016semi}, which iteratively approximates Chebyshev polynomials proposed by \cite{defferrard2016convolutional}, has motivated numerous novel designs. Some typical GNNs are reviewed below. 

\noindent \textbf{Graph Neural Networks.} GNNs generalize neural network into graph-structured data \cite{scarselli2008graph,kipf2016semi,klicpera2019predict,velivckovic2017graph}. The key operation is graph convolution, where information is routed from each node its neighbors with some deterministic rules (e.g., adjacency matrices and Laplacian matrices). For example, the propagation rule of GCN \cite{kipf2016semi} could be formulated as $\mathbf{H}^{(l+1)} = \sigma(\hat{\mathbf{A}} \mathbf{H}^{(l)} \mathbf{W}^{(l)})$, where $\hat{\mathbf{A}}$ denotes the normalized adjacency matrix with self-loop, $\sigma(.)$ denotes the non-linearity function, and $\mathbf{W}^{(l)}$ and $\mathbf{H}^{(l)}$is the learnable parameters and node representations at $l^{th}$ layer respectively. That of APPNP \cite{klicpera2019predict} is formulated as $\mathbf{Z}^{(k+1)} = (1 - \alpha) \hat{\mathbf{A}} \mathbf{Z}^{(k)} + \alpha \mathbf{H}$, where $\alpha$ denotes the teleport probabilities, $\mathbf{H}$ is the predicted class distribution before propagation and $\mathbf{Z}^{(k)}$ denotes the propagated class distribution at $k^{th}$ layer.

\noindent \textbf{From Graph Spectral Filter to Graph Neural Network.} The idea behind graph spectral filtering is modulating the graph signals with learnable parameters so that signals at different frequencies are either amplified or diminished. \cite{defferrard2016convolutional} proposes signal modulating with Chebyshev polynomials, which allows the model to learn neighbor information within K-hops with a relatively scalable amount of parameters. GCN proposed by \cite{kipf2016semi} approximates the K-th order Chebyshev polynomials by K convolution layers connected back-to-back, each of which assumes K equals to 1 and the maximal eigenvalue of Laplacian matrix equals to 2. Spatially, each propagation layer could be understood as gathering information from direct neighbors by mean pooling, and information from high-order neighbors could be captured through multiple stacked layers. Our proposed AKGNN simplifies this complexity of such filtering process by decoupling it into two sub-tasks: 1) limiting the scope of filters by finding the optimal trade-off between high and low frequency signals and 2) graph signal filtering on the distilled signals. 

\begin{figure*}[!h]
\centering
\includegraphics[width=1\linewidth]{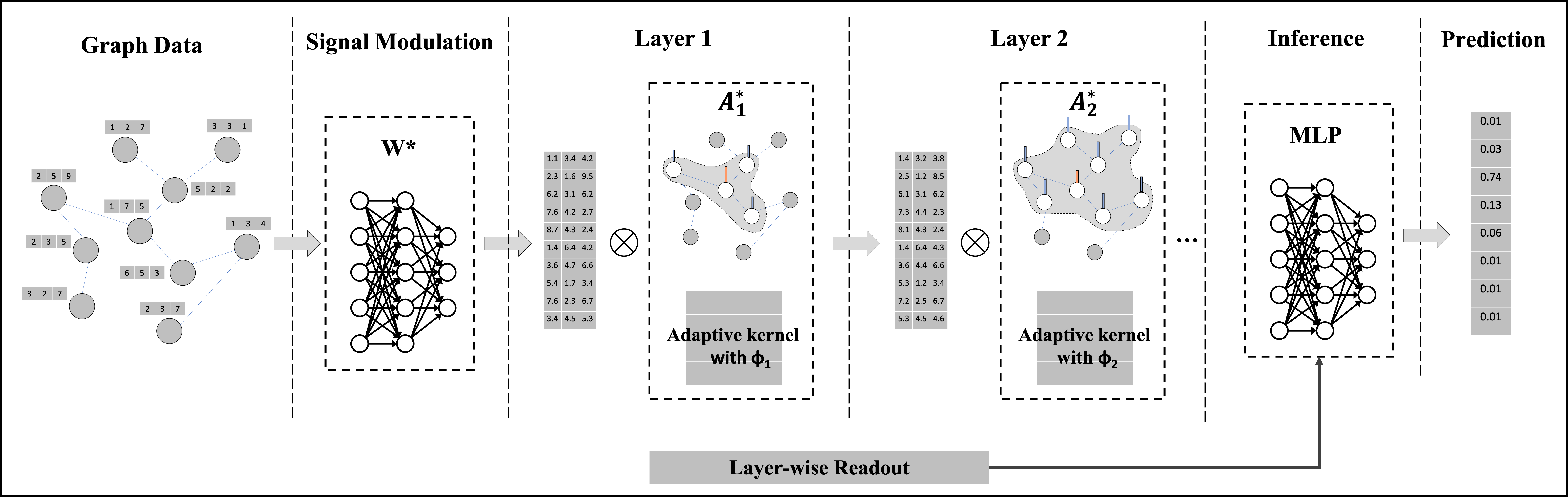}
\caption{AKGNN for node classification. The parameters of filters at all layers are approximated by a single MLP. And at each propagation layer, AKGNN learns the optimal trade-off between all-pass and low-pass filters and constructs $\mathbf{A}_{k}^{*}$ to conduct graph convolution. The class label is inferred by summing node representations at all layers through a prediction MLP.}
\label{system}
\end{figure*}

\section{Methodology}

In this section, we explain the technical details of Adaptive Kernel Graph Neural Network (AKGNN), as shown in Fig. \ref{system}. We present a type of graph convolution that is able to adaptively tune the weights of all-pass and low-pass filters by learning the maximal eigenvalue of graph Laplacian at each layer. Through such a design, the threshold between high and low frequency signals is efficiently optimized. Demonstrated by comprehensive experiments, proposed AKGNN is able to achieve state-of-the-art results on benchmarks acknowledged as community conventions.

\subsection{Adaptive Graph Kernel Learning}
\label{eigen}
Given an input graph $G$ and its normalized Laplacian matrix $\mathbf{L} = \mathbf{I} - \mathbf{D}^{-\frac{1}{2}}\mathbf{A}\mathbf{D}^{-\frac{1}{2}} = \mathbf{U}\mathbf{\Lambda}\mathbf{U}^T$, where $\mathbf{U}$ is the eigenvector matrix and $\mathbf{\Lambda}$ is the diagonal eigenvalue matrix of $\mathbf{L}$, Cheby-Filter \cite{defferrard2016convolutional} on a graph signal $\mathbf{f}$ is formulated as:

\begin{equation}
    \mathbf{f}^{'} = \sum_{k = 0}^{K} \mathbf{\theta}_k \mathbf{U} T_k(\tilde{\mathbf{\Lambda}}) \mathbf{U}^T \mathbf{f} 
    = \sum_{k = 0}^{K} \mathbf{\theta}_k T_k(\tilde{\mathbf{L}}) \mathbf{f},
    \label{cheb_poly}
\end{equation}
where $\mathbf{f}^{'}$ is the resulted modulated signal, $K$ is the order of the truncated polynomials, $\mathbf{\theta}_k$ denotes the learnable filter at $k^{th}$ order, $T_k(.)$ refers to the $k^{th}$ polynomial bases, $\tilde{\mathbf{\Lambda}}$ denotes the normalized diagonal eigenvalue matrix, and $\tilde{\mathbf{L}} = \mathbf{U} \tilde{\mathbf{\Lambda}} \mathbf{U}^T$.  $\tilde{\mathbf{\Lambda}} = \frac{2 \cdot \mathbf{\Lambda}}{\lambda_{max}} - \mathbf{I}$, where $\lambda_{max}$ denotes the maximum value in $\mathbf{\Lambda}$, has domain in range [-1,1]. Normalized form is used here instead of $\mathbf{\Lambda}$ since Chebyshev polynomial is orthogonal only in the range [-1,1] \cite{defferrard2016convolutional}. 

GCN \cite{kipf2016semi} simplifies the Cheby-Filter by assuming $K = 1$ and $\lambda_{max} \approx 2$ at each layer. Although the efficacy of GCN is promising, through this simplification, one issue is brought: graph convolution operation essentially conducts low-frequency filtering as proved by \cite{zhu2021interpreting}, where the similarity between adjacent nodes are enlarged as the number of propagation layers increases, and the kernel could not be adjusted to dataset where high-frequency information is important. Given that $T_0(\tilde{\mathbf{L}}) = \mathbf{I}$ and $T_1(\tilde{\mathbf{L}}) = \tilde{\mathbf{L}}$ \cite{defferrard2016convolutional}, Eq. \ref{cheb_poly} is re-formulated as follows:

\begin{equation} \label{our_eq1}
\begin{split}
\mathbf{f}^{'} &= \mathbf{\theta}_0 \mathbf{I} \mathbf{f} + \mathbf{\theta}_1 (\frac{2}{\lambda_{max}}(\mathbf{I} - \mathbf{D}^{-\frac{1}{2}}\mathbf{A}\mathbf{D}^{-\frac{1}{2}}) - \mathbf{I}) \mathbf{f}\\
               &= \mathbf{\theta}_0 \mathbf{I} \mathbf{f} + \frac{2}{\lambda_{max}} \mathbf{\theta}_1 \mathbf{I} \mathbf{f} -
                  \frac{2}{\lambda_{max}} \mathbf{\theta}_1 \mathbf{D}^{-\frac{1}{2}}\mathbf{A}\mathbf{D}^{-\frac{1}{2}} \mathbf{f} - \mathbf{\theta}_1 \mathbf{I} \mathbf{f}.
\end{split}
\end{equation}

By setting $\mathbf{\theta}_0 = -\mathbf{\theta}_1$, Eq. \ref{our_eq1} can be simplified as follows:

\begin{equation} \label{our_eq2}
\begin{split}
\mathbf{f}^{'} &= \mathbf{\theta}_0 \mathbf{I} \mathbf{f} + \frac{2}{\lambda_{max}} \mathbf{\theta}_1 \mathbf{I} \mathbf{f} -
                  \frac{2}{\lambda_{max}} \mathbf{\theta}_1 \mathbf{D}^{-\frac{1}{2}}\mathbf{A}\mathbf{D}^{-\frac{1}{2}} \mathbf{f} - \mathbf{\theta}_1 \mathbf{I} \mathbf{f}\\
               &= \mathbf{\theta}_0 \mathbf{I} \mathbf{f} - \frac{2}{\lambda_{max}} \mathbf{\theta}_0 \mathbf{I} \mathbf{f} +
                  \frac{2}{\lambda_{max}} \mathbf{\theta}_0 \mathbf{D}^{-\frac{1}{2}}\mathbf{A}\mathbf{D}^{-\frac{1}{2}} \mathbf{f} + \mathbf{\theta}_0 \mathbf{I} \mathbf{f} \\
               &= \frac{2\lambda_{max}-2}{\lambda_{max}} \mathbf{\theta}_0 \mathbf{I} \mathbf{f} + \frac{2}{\lambda_{max}} \mathbf{\theta}_0 \mathbf{D}^{-\frac{1}{2}}\mathbf{A}\mathbf{D}^{-\frac{1}{2}} \mathbf{f}
\end{split}
\end{equation}
In Eq. \ref{our_eq2}, the first and second term could be regarded as the all-pass filter with weight  $\frac{2\lambda_{max}-2}{\lambda_{max}}$ and low-pass filter with weight $\frac{2}{\lambda_{max}}$, respectively. With these settings, we have the following theorem with theoretical analysis later. 

\textbf{Theorem 1.} \textit{While conducting spectral modulation, we can control the balance between all-pass and low-pass filters by simply tuning $\lambda_{max}$. And $\lim_{\lambda_{max}\to\infty} \mathbf{f}^{'} \approx \mathbf{\theta}_0 \mathbf{I} \mathbf{f}$, which is an all-pass filter; $\lim_{\lambda_{max}\to1} \mathbf{f}^{'} \approx \mathbf{\theta}_0 \mathbf{D}^{-\frac{1}{2}}\mathbf{A}\mathbf{D}^{-\frac{1}{2}} \mathbf{f}$, which is a low-pass filter.}

We can find the optimal threshold between low and high frequency signals by tuning the weights of these two filters. With a high $\lambda_{max}$, the weight of all-pass filters is elevated and so are the high frequency signals. Whereas when $\lambda_{max}$ is low, the weight of low-pass filters is high and so are the low frequency signals. However, it is nontrivial to manually decide which part is more significant than the other as situations are different across various datasets. A natural step forward would be finding the optimal eigenvalues in a data-driven fashion. Hence, in order to find the optimal threshold, we make $\lambda_{max}$ at $k^{th}$ layer a learnable parameter:

\begin{equation}
    \lambda_{max}^{k} = 1 + \text{relu}(\phi_k),
    \label{lambda}
\end{equation}
where $\phi_k \in \mathbb{R}$ is a learnable scalar, and relu(.) refers to rectified linear unit function. $\phi_k$ is initialized as 1, since $\lambda_{max}^{k} = 2$ when $\phi_k = 1$. Under this setting, the initial balance between two filters is identical (i.e., $\frac{2\lambda_{max}-2}{\lambda_{max}} = \frac{2}{\lambda_{max}} = 1$), preventing the model from being stuck at local-minimum. We regularize $\phi_k$ by a relu function because relu has a codomain of [0, $\infty$]. This enables the propagation layer to achieve a all-pass filter when $\phi_k \rightarrow \infty$ or a low-pass filter when  $\phi_k \rightarrow 0$. Utilizing a layer-wise matrix representation, we have node embedding $\mathbf{H}^{(k)}$ at $k^{th}$ layer as:

\begin{equation} \label{our_eq3}
\begin{split}
&\mathbf{H}^{(k)} = (\frac{2\lambda^k_{max}-2}{\lambda^k_{max}}\mathbf{I} +  \frac{2}{\lambda^k_{max}} \mathbf{D}^{-\frac{1}{2}}\mathbf{A}\mathbf{D}^{-\frac{1}{2}}) \mathbf{H}^{(k-1)} \mathbf{W}_k \\
                 &= (\frac{2\text{relu}(\phi_k)}{1+\text{relu}(\phi_k)} \mathbf{I} + \frac{2}{1+\text{relu}(\phi_k)} \mathbf{D}^{-\frac{1}{2}}\mathbf{A}\mathbf{D}^{-\frac{1}{2}}) \mathbf{H}^{(k-1)} \mathbf{W}_k,
\end{split}
\end{equation}
where $\mathbf{H}^{(0)} = \mathbf{X}$, $\mathbf{W}_k \in \mathbb{R}^{d^{(k-1)} \times d^{(k)}}$ denotes the parameter matrix of filter at $k^{th}$ layer and $d^{(k)}$ refers to the dimension of signals at $k^{th}$ layer. The domain of eigenvalues of $(\frac{2\text{relu}(\phi_k)}{1+\text{relu}(\phi_k)} \mathbf{I} + \frac{2}{1+\text{relu}(\phi_k)} \mathbf{D}^{-\frac{1}{2}}\mathbf{A}\mathbf{D}^{-\frac{1}{2}})$ is [0, 2], which can introduce numerical instabilities and unexpected gradient issues. So using the renormalization trick proposed in \cite{kipf2016semi}, we further normalize and reformat $(\frac{2\text{relu}(\phi_k)}{1+\text{relu}(\phi_k)} \mathbf{I} + \frac{2}{1+\text{relu}(\phi_k)} \mathbf{D}^{-\frac{1}{2}}\mathbf{A}\mathbf{D}^{-\frac{1}{2}})$ as $\mathbf{A}_{k}^{*} = \mathbf{D}_k^{-\frac{1}{2}}\mathbf{A}_k\mathbf{D}_k^{-\frac{1}{2}}$, where $\mathbf{A}_k = \frac{2\text{relu}(\phi_k)}{1+\text{relu}(\phi_k)} \mathbf{I} + \frac{2}{1+\text{relu}(\phi_k)} \mathbf{A}$, and $\mathbf{D}_k$ denotes the diagonal degree matrix of $\mathbf{A}_k$. Putting them together, the layer-wise propagation is summarized as:

\begin{equation}
    \mathbf{H}^{(k)} = \mathbf{A}_{k}^{*} \mathbf{H}^{(k-1)} \mathbf{W}_k.
    \label{propagate_1}
\end{equation}

Many current GNNs \cite{klicpera2019predict,zhu2009introduction,chen2020simple} have adopted kernels where the balance between all-pass and low-pass filters are dedicatedly tailored. They utilize a pre-defined balancing variable to achieve so but finding the optimal threshold for a specific dataset is undeniably non-trivial as the search space is usually very large. Different from these current approaches, the adjacency matrix $\mathbf{A}_{k}^{*}$ we utilize at $k^{th}$ layer is parameterized with a single scalar. This design enables our model to effectively learn the optimal balance between high and low frequency signals during the training phase and omit the cumbersome hyper-parameter tuning. However, it is difficult for the model to simultaneously learn both the filter $\mathbf{W}_k$ and the optimized graph Laplacian $\mathbf{A}_{k}^{*}$ since the filter operates on the current version of graph Laplacian and dynamically updating both might lead to a situation where the whole model will never converge. Moreover, as we stack numerous layers to capture the high-order information, $\mathbf{W}_k$ still introduces a number of parameters, which are very likely to introduce the over-fit issue. Hence we utilize a parameterization trick to alleviate the above issues.  

\subsection{Parameterization trick}

The key motivation of all graph convolutions to stack multiple propagation layers is capturing high-order information that is beneficial to downstream tasks. As mentioned in the introduction, aiming to capture such information, under the designs of most GNNs, more parameters are also introduced (e.g., $\mathbf{W}_k$ in Eq. \ref{propagate_1}). This could bring the over-fitting problem when nodes are scarcely labeled and offset the intended benefits. \cite{wu2019simplifying} proposes to approximate parameters at all layers with a single matrix and meanwhile eliminate the non-linearity in between, which is proved to achieve similar results with fewer parameters. Nevertheless, by conducting such approximation, the complexity of the graph filter is also significantly decreased, making dynamically tuning both the filter and graph Laplacian feasible. Hence, we utilize a modified version of such parameterization trick to approximate parameters at all layers with a single matrix. Specifically, we can re-write Eq. \ref{propagate_1} by expanding $\mathbf{H}^{(k-1)}$ as follows:

\begin{equation} \label{propagate_2_1}
\begin{split}
    \mathbf{H}^{(K)} &= \mathbf{A}_{k}^{*} \mathbf{H}^{(K-1)} \mathbf{W}_K \\
                     &= \mathbf{A}_{K}^{*} \mathbf{A}_{K-1}^{*} \dots \mathbf{A}_{1}^{*} \mathbf{X} \mathbf{W}_1 \dots \mathbf{W}_{K-1} \mathbf{W}_K,
\end{split}
\end{equation}
where $K$ is the total number of propagation layers. We propose to use a single matrix $\mathbf{W}^{*} \in \mathbb{R}^{d \times d^{(K)}}$ to approximate the functionalities of all $\mathbf{W}_k$, such that:
\begin{equation}
\begin{split}
    \mathbf{H}^{(k)} = \mathbf{A}_{k}^{*} \mathbf{H}^{(k-1)}\; \; \; \;  \text{for } k \geq 1, \text{and }\mathbf{H}^{(0)} = \sigma(\mathbf{X} \mathbf{W}^{*}),
    \label{propagate_2}
\end{split}
\end{equation}
where $\sigma(.)$ denotes the ReLU non-linearity. From the perspective of spatial aggregation, intuitively, $i^{th}$ row of $\mathbf{H}^{(k)}$ is simply a linear combination of the modulated graph signals of node $v_i$ and those of its neighbors, whose distances to $v_i$ are within $k$ hops. Through this trick, each convolution layer has only one parameter $\phi_k$, which also significantly alleviates the convergence issue.

\subsection{Inference and Prediction}
\label{inference}
After performing signal modulation for $K$ propagation layers, we generate $K$ node representation matrices $\{\mathbf{H}^{(k)}|1\leq k \leq K\}$. These matrices are combined through a readout function and then fed into a Multi-layer Perceptron (MLP) to predict the class labels $\mathbf{Y}^P \subseteq \{0,1\}^{N \times C} $, formulated as:

\begin{equation}
    \mathbf{Y}^P = \text{softmax}(f_{MLP}(READOUT(\mathbf{H}^{(k)}))),
\end{equation}
where $READOUT(.)$ denotes the layer-wise readout function. We choose to combine intermediate node representations through a readout function, instead of using $\mathbf{H}^{(k)}$ directly for the final prediction, because it is very possible that different nodes require distinct levels of information for node classification. And bringing high-order information for nodes whose labels could be inferred merely through local information might jeopardize the decision margin \cite{xu2018representation,liu2020towards}. As for the selection of readout function, we explore element-wise summation $sum(.)$ instead of other element-wise operations (e.g., mean, or max) to maximize the express power \cite{xu2018powerful}. Compared with other readout functions, the summation function is injective and with such function, we reduce the possibility of nodes, that share the same graph signal coefficients but are structurally different, being represented the same. Layer-wise concatenation is also able to achieve similar functionalities but it also introduces the number of parameters in $f_{MLP}(.)$. 

\subsection{Theoretical Analysis}
\label{eigen_proof}
In this section, we explain the mechanism of adaptive kernel learning with maximal eigenvalue from the perspective of spectral graph theory. Recall that graph Laplacian is formulated as $\mathbf{L} = \mathbf{D} - \mathbf{A}$ or its normalized form  $\mathbf{L} = \mathbf{D}^{-\frac{1}{2}}(\mathbf{D} - \mathbf{A})\mathbf{D}^{-\frac{1}{2}} = \mathbf{I} - \mathbf{D}^{-\frac{1}{2}}\mathbf{A}\mathbf{D}^{-\frac{1}{2}}$. The Laplacian matrix of either form is symmetric and semi-positive definite, which gives it an important property: it can be eigen-decomposed such one set of resulted eigenvalues are all greater than or equal to zero, formulated as $\mathbf{L} = \mathbf{U} \mathbf{\Lambda} \mathbf{U}^T$. We sort the eigenvalues and their corresponding eigenvectors such that $0 = \lambda_1 \leq \lambda_2 \dots \leq \lambda_{N-1} \leq \lambda_N$. The key idea behind graph spectral filtering is modulating the graph signals' frequencies so that these beneficial to downstream tasks are magnified while others are diminished. This could be achieved by modifying the eigenvalues of Laplacian matrix with a filter function $f(.)$. \cite{defferrard2016convolutional} proposes to utilize model $f(.)$ by Chebyshev polynomials $T_k(.)$. In the work of \cite{defferrard2016convolutional}, it is prerequisite that polynomials at different orders are orthogonal, because orthogonality guarantees that modifying filter at a specific order won't interfere with other orders. So it is necessary to normalize $\mathbf{\Lambda}$ as $\tilde{\mathbf{\Lambda}} = \frac{2 \cdot \mathbf{\Lambda}}{\lambda_{max}} - \mathbf{I}$ such that its domain aligns with the domain [-1, 1] where the orthogonality of Chebyshev polynomials is defined. Under this setup, in order to modulate signals, at least $\mathcal{O}(d \times K)$ parameters are needed, where $d, K$ stands for the dimension of input signals and the order of truncated polynomials, respectively. To reduce the complexity of this process, we propose to make $\lambda_{max}$ as a learnable parameter. $\lambda_{max}$ as a single parameter could effectively solve one major task of the graph filter: balancing the trade-off between high and low frequencies. When $\lambda_{max}$ is large (e.g., $\lim_{\lambda_{max}\to\infty}$), all values on the diagonal of $\mathbf{\Lambda}$ becomes infinitely close to each other and hence every signal in the original graph Laplacian is retained, corresponding to the all-pass filter $\lim_{\lambda_{max}\to\infty} \mathbf{f}^{'} \approx \mathbf{\theta}_0 \mathbf{I} \mathbf{f}$ in Theorem 1. Notice that the domain of normalized Laplacian matrix is upper-bounded by 2 and we allow the maximum eigenvalue to freely vary between (1, $\inf$). In this case, our goal here is not to find the actual maximum eigenvalue; instead, we aim to utilize this normalization process such that the trade-off between all-pass and low-pass filters is optimized. If $\lambda_{max}$ is upper-bounded by 2, in circumstances where high-frequency signals are significant for downstream tasks, the best scenario we can possibly approach is the equal weight on all-pass and low-pass filters (i.e., $\frac{2\lambda_{max}-2}{\lambda_{max}} \mathbf{\theta}_0 \mathbf{I} \mathbf{f} + \frac{2}{\lambda_{max}} \mathbf{\theta}_0 \mathbf{D}^{-\frac{1}{2}}\mathbf{A}\mathbf{D}^{-\frac{1}{2}} \mathbf{f} = \mathbf{I} \mathbf{f} + \mathbf{\theta}_0 \mathbf{D}^{-\frac{1}{2}}\mathbf{A}\mathbf{D}^{-\frac{1}{2}} \mathbf{f}$). 

On the other hand, when $\lambda_{max}$ is small (e.g., $\lim_{\lambda_{max}\to 1}$), we will have some high frequency signals whose eigenvalues are larger than $\lambda_{max}$. In this case, these signals will become unorthogonal to low frequency signals whose eigenvalues are less than or equal to $\lambda_{max}$ and these high frequency signals can be generated by linear combinations of the low ones, which corresponds to low-pass filter $\lim_{\lambda_{max}\to1} \mathbf{f}^{'} \approx \mathbf{\theta}_0 \mathbf{D}^{-\frac{1}{2}}\mathbf{A}\mathbf{D}^{-\frac{1}{2}} \mathbf{f}$. With the help of learnable $\lambda_{max}$, the complexity of graph signal filtering in AKGNN is significantly reduced. Because the scope of signal sets is diminished by learnable $\lambda_{max}$ and filter only needs to focus on signals beneficial to the downstream task.

\subsection{Complexity Analysis}

The complexity of AKGNN can be decoupled into two parts. The first portion is graph signal filtering with $f_{MLP}(.)$, with complexity $\mathcal{O}(Nd\cdot d^{(K)})$, where N stands for the number of nodes, $d$ refers to the dimension of the input signal, and $d^{(K)}$ is the dimension of the filtered signal. The second portion is graph convolution with the adaptive kernel, with complexity $\mathcal{O}(|E|d^{(K)})$ per each layer, where $|E|$ denotes the number of edges. Hence for a $K$-layer AKGNN, the total computational complexity is $\mathcal{O}(N\cdot d\cdot d^{(K)} + K\cdot|E|\cdot d^{(K)})$, which is linear with the number of nodes, edges, and layers. 

\section{Experiments and Analysis}

We follow the experiment setup acknowledged as community conventions, the same as node classification tasks in \cite{kipf2016semi,velivckovic2017graph} (e.g., publicly fixed 20 nodes per class for training, 500 nodes for validation, and 1,000 nodes for testing). The three datasets we evaluate are Cora, Citeseer and Pubmed \cite{sen2008collective}. 

\noindent \textbf{Baselines.} We compare AKGNN with three GNN branches:
\begin{itemize}
    \item \textbf{\textit{Graph convolutions}}: ChebyNet \cite{defferrard2016convolutional}, GCN \cite{kipf2016semi}, GAT \cite{velivckovic2017graph}, JKNets \cite{xu2018representation}, APPNP \cite{klicpera2019predict}, SGC \cite{wu2019simplifying} and SSGC \cite{zhu2021simple}.
    \item \textbf{\textit{Regularization based}}: VBAT \cite{deng2019batch}, Dropedge \cite{rong2020dropedge}, GAugO \cite{zhao2020data}, and PGSO \cite{dasoulas2021learning}. The backbone model used is GCN.
    \item \textbf{\textit{Sampling based}}: GraphSage \cite{hamilton2017inductive}, FastGCN \cite{chen2018fastgcn}.
\end{itemize}

\subsubsection{Implementation Detail}
We utilize PyTorch as our deep learning framework to implement AKGNN. The adaptive kernel learning mechanism is engineered in a sparse tensor fashion for compact memory consumption and fast back-propagation. The weights are initialized with Glorot normal initializer \cite{glorot2010understanding}. We explore Adam to optimize parameters of AKGNN with weight decay and use early stopping to control the training iterations based on validation loss. Besides, we also utilize dropout mechanism between all propagation layers. All the experiments in this work are implemented on a single NVIDIA GeForce RTX 2080 Ti with 11 GB memory size and we didn't encounter any memory bottleneck issue while running all experiments. 

\subsubsection{Hyperparameter Detail}
\label{setting}
In AKGNN, related hyperparameters are the number of layers $K$, hidden size $d^{(K)}$, dropout rate, learning rate, weight decay rate, and patience for early stopping. We utilize identical hyperparameter settings over all three datasets as our model learns to adapt to different dataset. The number of layers $K$ is 5, the hidden size $d^{(K)}$ is 64, the dropout rate between propagation layers is 0.6, the learning rate is 0.01, the weight decay rate is 5e-4, and patience for early stopping is 100 iterations. 

\begin{table}
\centering
\begin{tabular}{c|ccc}
    \toprule
    \toprule
    \multirow{3}{*}{Method} & \multicolumn{3}{c}{Graph} \\
    \cmidrule(r){2-4}
         & Cora     & Citeseer & Pubmed \\
    \midrule
    ChebyNet  & 81.2       & 69.8            & 74.4\\
    GCN       & 81.5       & 70.3            & 79.0 \\
    GAT       & 83.0$\pm$0.7 & 72.5$\pm$0.7  & 79.0$\pm$0.3 \\
    APPNP     & 83.8$\pm$0.3 & 71.6$\pm$0.5  & 79.7$\pm$0.3 \\
    SGC       & 81.0$\pm$0.0 & 71.9$\pm$0.1  & 78.9$\pm$0.0 \\
    SSGC      & 83.5$\pm$0.0 & 73.0$\pm$0.1  & 80.2$\pm$0.0 \\
    JKNets    & 83.3$\pm$0.5 & 72.6$\pm$0.3  & 79.2$\pm$0.3 \\

    \midrule
    PGSO      & 82.5$\pm$0.3 & 71.8$\pm$0.2  & 79.3$\pm$0.5 \\
    VBAT      & 83.6$\pm$0.5 & 73.1$\pm$0.6  & 79.9$\pm$0.4 \\
    GAugO     & 83.6$\pm$0.5 & 73.3$\pm$1.1 & 80.2$\pm$0.3 \\
    Dropedge  & 82.8         & 72.3          & 79.6         \\
    \midrule
    GraphSage & 78.9$\pm$0.6 & 67.4$\pm$0.7  & 77.8$\pm$0.6 \\
    FastGCN   & 81.4$\pm$0.5 & 68.8$\pm$0.9  & 77.6$\pm$0.5 \\
    \midrule
    \textbf{AKGNN} (ours)   & \textbf{84.4$\pm$0.3} & \textbf{73.5$\pm$0.2} & \textbf{80.4$\pm$0.3} \\
    % \textbf{AKGNN} (rand. split)   & 86.2$\pm$0.8 & 75.3$\pm$0.4 & 82.5$\pm$0.7 \\
    \midrule
    \midrule
    w/o $\lambda$ learning  & 81.4$\pm$0.2 & 71.9$\pm$0.1 & 79.1$\pm$0.2 \\
    w/o PT         & 83.1$\pm$0.1 & 72.2$\pm$0.5 & 80.1$\pm$0.3 \\
    w/o readout    & 83.5$\pm$0.2 & 73.1$\pm$0.3 & 79.4$\pm$0.2 \\
    \bottomrule
    \bottomrule
  \end{tabular}
  \caption{Overall classification accuracy (\%).}
  \label{exp_overall}
 \end{table}

\subsection{Overall Results}

From the upper portion of Tab. \ref{exp_overall}, we observe that AKGNN consistently outperforms all baselines by a noticeable margin over all datasets. Comparing AKGNN with the best-performing baseline of each dataset, we further conduct t-test and the improvement margin is statistically significant with the p-values less than 0.001. The improvement of AKGNN over GCN is 3.2\%, 3.7\% and 1.4\% on Cora, Citeseer and Pubmed; whereas that over GAT is 1.4\%, 1.3\% and 1.4\%. Compared with JKNet that utilizes a similar global readout function and parameterization trick, AKGNN has 1.1\%, 0.9\% and 1.2\% improvements, demonstrating the efficacy of adaptive kernel learning mechanism. As the graph regularization-based model gets more attention, we also compare AKGNN with these recent works and the improvements are 0.8\%, 0.2\% and 0.2\%. 

\begin{figure}[h]
\centering
\includegraphics[width=1\linewidth]{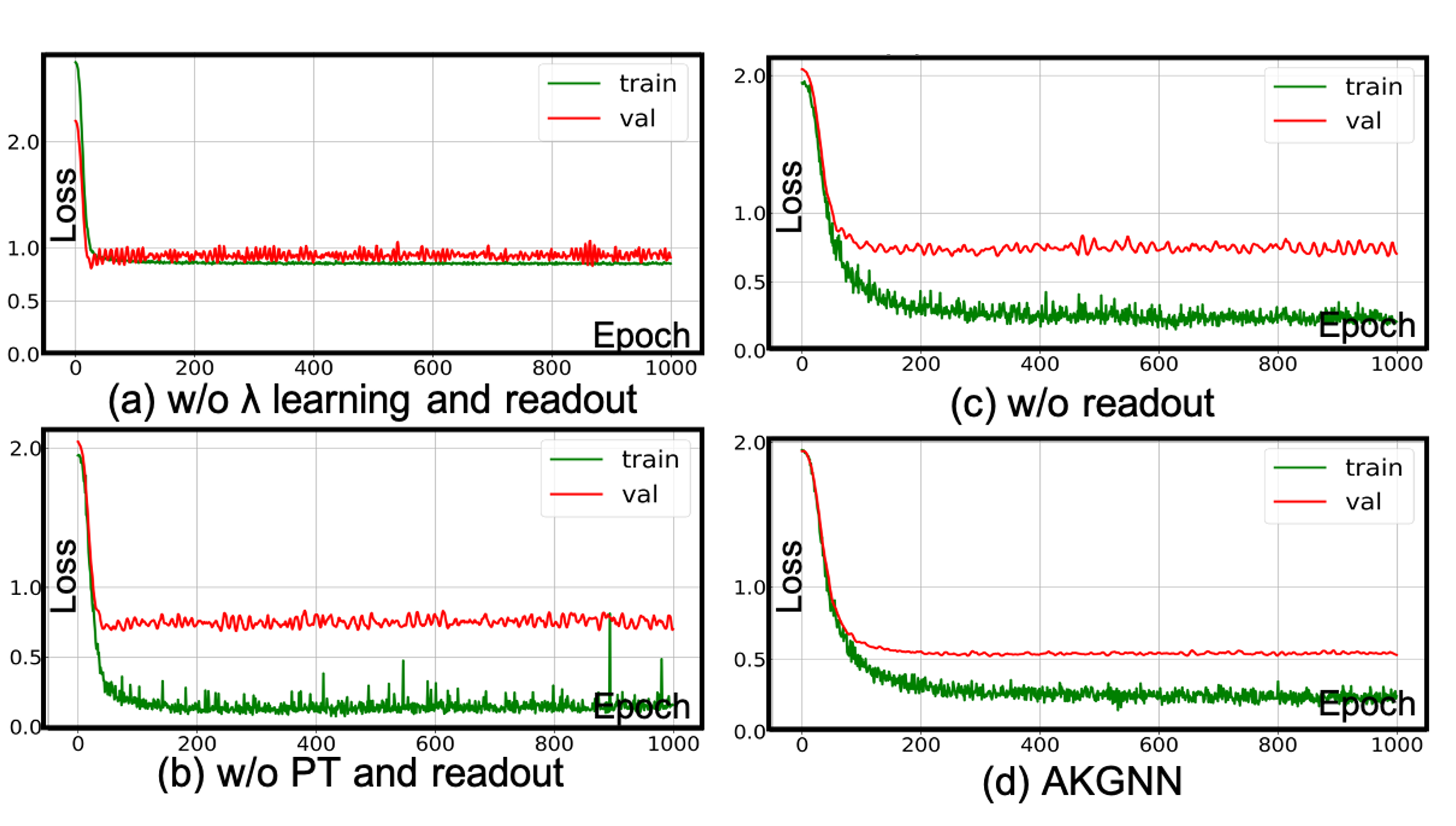}
\caption{Generalization on Cora.}
\label{loss_fig}
\end{figure}

\subsection{Ablation Studies}

To analyze the contribution of different components in AKGNN, we conduct several sets of ablation studies. In order to examine the effectiveness of $\lambda_{max}$ learning, we design the first variant as our model without adaptive kernel learning, denoted as `w/o $\lambda$'. Another variant is our model without parameterization trick, denoted as `w/o PT', aimed at validating its effectiveness in combating the over-fitting issue. The last variant is our model without readout function (i.e., $sum(.)$), in order to prove that nodes require different levels of information to achieve better performance. From the bottom portion of Tab. \ref{exp_overall}, we can first observe that all components contribute to the performance of AKGNN. The first variant without adaptive kernel learning and readout experiences a significant performance downgrade and is worse than the vanilla GCN model on some datasets, because we stack a lot of convolution layers and it encounters the over-smoothing. Comparing AKGNN without readout function with baselines, we observe similar performance. 

In Fig. \ref{loss_fig}.a, both training and validation loss of this variant are highest and the differences between them are smallest across all variants, indicating that the over-smoothing issue has caused node representations to be indistinguishable. The second variant without parameterization trick has the lowest training loss, and also the biggest gap between training and validation loss, as shown in Fig. \ref{loss_fig}.b. This phenomenon represents the model suffers from the over-fitting problem due to the large number of parameters brought by numerous propagation layers. The third variant without readout function relatively performs better than the previous two, but still not as good as AKGNN, as shown in Fig. \ref{loss_fig}.c. This illustrates that the decision boundary is diminished as a result of bringing redundant high-order information for nodes that requires only low-frequency signals. Nevertheless, we further examine the generalization ability of our proposed method. As shown in Fig. \ref{loss_fig}.d, we can observe the lowest validation loss across all variants and meanwhile the differences between training and validation loss remain small, which demonstrates the generalization ability of AKGNN. 

\begin{figure}[h]
\centering
\includegraphics[width=1\linewidth]{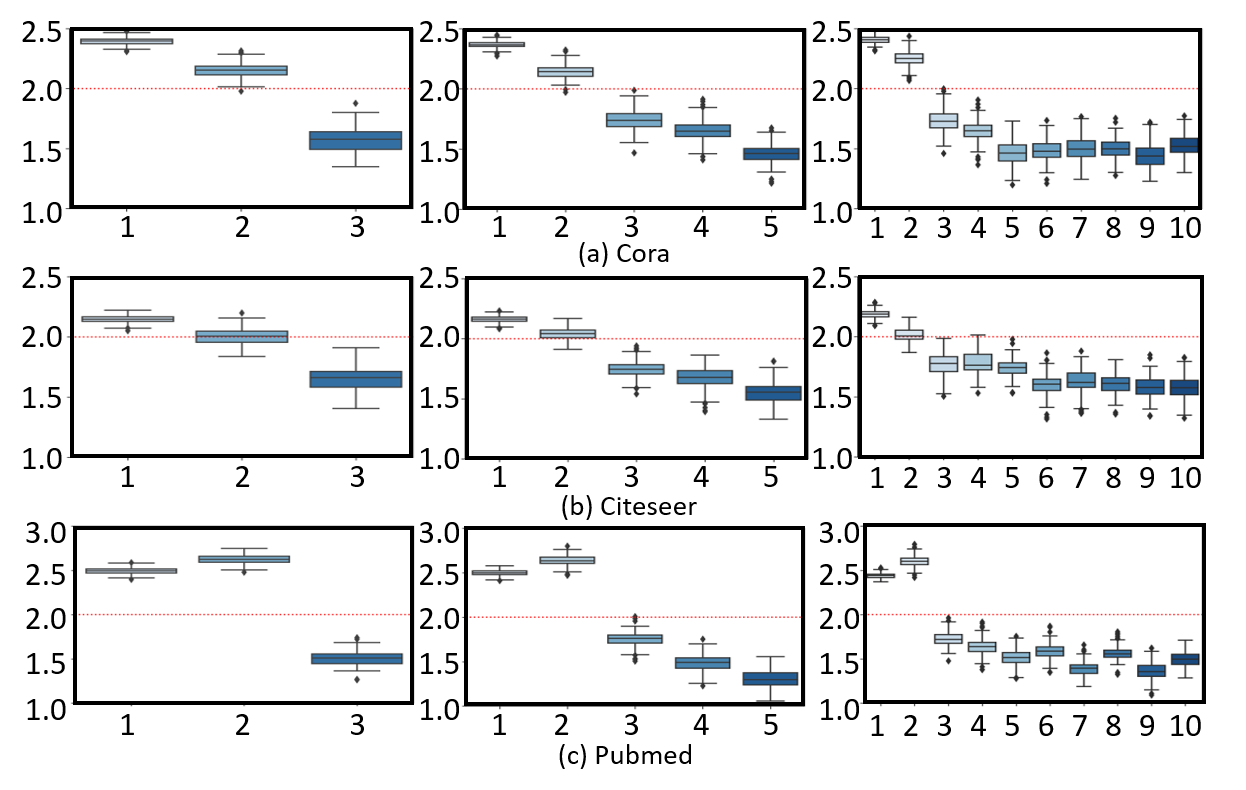}
\caption{Maximal eigenvalues vs. number of layers ($x$-axis: layer $k$, $y$-axis: $\lambda_{max}^k$). Red dashed line indicates the equal weights of all-pass and low-pass filters. A higher $\lambda_{max}^k$ indicates a higher weight of all-pass filter; whereas a low $\lambda_{max}^k$ indicates a higher weight of low-pass filter.}
\label{lambda_fig}
\end{figure}

\subsection{Analysis of Adaptive Kernel Learning}

The key motivation of learning the maximal eigenvalue is learning the optimal threshold between low and high frequency signals. In order to examine how it combats against the generality issue, we visualize the maximal eigenvalue at each layer, as shown in Fig. \ref{lambda_fig}. We first analyze the dynamics of maximal eigenvalues $\lambda_{max}^k$ within the same dataset. We can notice that the value of $\lambda_{max}^k$ incrementally decreases as $k$ progresses and reaches a plateau where $\lambda_{max}^k$ doesn't change much after fifth layer across all datasets. We interpret this phenomenon as our model enforcing high-order layers to become meaningless. Because a low maximal eigenvalue at high-order layer would make node representations getting more indistinguishable. Moreover, $\lambda_{max}^k$ at early layers doesn't deviate as $K$ increases, demonstrating the strong contribution of local information and stability of AKGNN. Then we analyze the dynamics between these three datasets. We can notice that the $\lambda_{max}^k$ of Pubmed has a higher mean than those of other two, showing that for node classification, high frequency signals benefits most for Pubmed dataset. Meanwhile, we can observe a significant $\lambda_{max}^k$ drop at the second layer for Pubmed, indicating the redundancy of high-order information. This phenomenon also aligns with GNNs like GCN or GAT performing best with only two layers. Besides an intuitive explanation given above, we also theoretically explicate these phenomenon. Adapted Laplacian matrices across all layers share the same eigenvectors $\mathbf{U}$, because essentially our operation only modifies the diagonal eigenvalue matrix $\mathbf{\Lambda}$. Hence the commutative rule for matrix multiplication applies for all $\mathbf{A}_{k}^{*}$ as they are simultaneously diagonalizable and the order of $\mathbf{A}_{k}^{*}$ multiplication in Eq. \ref{propagate_2_1} could be switch. In short, higher learned maximum eigenvalues should be observed if high-frequency signals dominate; whereas lower ones should be observed if low-frequency signals dominate. Across these three datasets, we can observe that AKGNN learns relatively high maximum eigenvalues compared with Cora and Pubmed, which aligns with the homophily property of there three datasets (i.e., Citeseer has the lowest homophily ratio.).

\subsection{Analysis of Number of Propagation Layers}

\begin{figure}[h]
\centering
\includegraphics[width=1\linewidth]{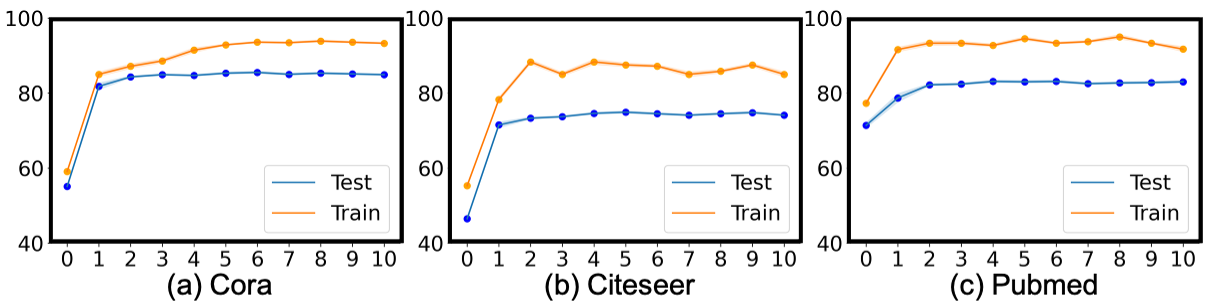}
\caption{Impact of the number of layers on accuracy. ($x$-axis: $K$, $y$-axis: accuracy (\%))}
\label{layer_fig}
\end{figure}

Beside adaptation of global readout function that has been utilized in \cite{liu2020towards, xu2018representation}, our adaptive kernel learning mechanism also improves AKGNN's resilience to the over-reaching by enforcing high-order layers to become meaningless, as discussed in the previous sections. And the impact of the number of layers on performance is shown in Fig. \ref{layer_fig}. From this figure, we can notice that the accuracy on both testing and training reaches their highest around fifth layer and remains stable as the number of layers increases, demonstrating the AKGNN's strong resilience to the over-smoothing while the model is the over-reaching. We don't conduct experiments on AKGNN with more than 10 layers because 10-hop sub-graph of a target node covers almost all its possibly reachable nodes; the resilience to the over-fitting of AKGNN is only a byproduct of adaptive kernel learning and not the focus of this work. 
\section{Conclusion}
In this work, we study the problem of node representation learning on graphs and present Adaptive Kernel Graph Neural Network (AKGNN). In AKGNN, we propose adaptive kernel learning to find the optimal threshold between high and low frequency signals. Together with parameterization trick and global readout function, AKGNN is highly scalable, achieves competitive performance and retains so even with a number of convolution layers. Through experiments on three acknowledged benchmark datasets, AKGNN outperforms all baselines. Different from other graph convolution models whose guiding kernels are fixed and not ideal to all kinds of graphs, AKGNN learns to adapt to different graph Laplacians, which could shed light on a different path while researchers design new GNN models. We do not observe ethical concern or negative societal impact entailed by our method. However, care must be taken to ensure positive and societal consequences of machine learning. In the future, we aim to transfer the similar ideas of AKGNN to directed graphs or investigating the possibility of applying adaptive kernel learning to other kernels.

\pagebreak

\section{Acknowledgement}
This work is partially supported by the NSF under grants IIS-2107172, IIS-2140785, CNS-1940859, CNS-1814825, IIS-2027127, IIS-2040144, IIS-1951504 and OAC-1940855, the NIJ 2018-75-CX-0032.

\bibliography{aaai22}
\end{document}